%% file: acl_latex.tex
\pdfoutput=1
\documentclass[11pt]{article}

\usepackage[]{acl}

\usepackage{times}
\usepackage{latexsym}
\usepackage{mathbbol}
\usepackage{amsmath}
\usepackage{graphicx}
\usepackage{multirow}
\DeclareMathOperator{\softmax}{softmax}

\usepackage[T1]{fontenc}
\usepackage[utf8]{inputenc}

\usepackage{microtype}

\usepackage{inconsolata}

\title{Incorporating Precedents for Legal Judgement Prediction \\ on European Court of Human Rights Cases}

\author{ Santosh T.Y.S.S, \bf{Mohamed Hesham Elganayni,} \\ \bf{Stanisław Sójka, Matthias Grabmair} \\ School of Computation, Information, and Technology; \\
Technical University of Munich, Germany \\ \ }

\begin{document}
\maketitle
\begin{abstract}
Inspired by the legal doctrine of stare decisis, which leverages precedents (prior cases) for informed decision-making, we explore methods to integrate them into LJP models. To facilitate precedent retrieval, we train a retriever with a fine-grained relevance signal based on the overlap ratio of alleged articles between cases. We investigate two strategies to integrate precedents: direct incorporation at inference via label interpolation based on case proximity and during training via a precedent fusion module using a stacked-cross attention model. We employ joint training of the retriever and LJP models to address latent space divergence between them. Our experiments on LJP tasks from the ECHR jurisdiction reveal that integrating precedents during training coupled with joint training of the retriever and LJP model, outperforms models without precedents or with precedents incorporated only at inference, particularly benefiting sparser articles.

\end{abstract}

\section{Introduction}
\input{text/Introduction}

\section{Method: Incorporating Precedents}
\input{text/Method}

\section{Experiments \& Results}
\input{text/Experiments}

\section{Conclusion}
\input{text/conclusion}

\section*{Limitations}
\input{text/limitations}

\section*{Ethics Statement}
\input{text/ethics}

\bibliography{custom}

\appendix
\input{text/appendix}

\end{document}

%% file: text/Introduction.tex
The task of case outcome classification deals with identifying the outcome from a textual description of case facts and is generally referred to as Legal Judgement Prediction (LJP) (e.g., \citealt{aletras2016predicting, chalkidis2019neural}). It has been studied using corpora from different jurisdictions, such as the European Court of Human Rights (ECHR) \cite{chalkidis2019neural, chalkidis2021paragraph, chalkidis2022lexglue, aletras2016predicting, medvedeva2020using, santosh2022deconfounding,santosh2023Zero},
%says2020prediction, liu2017predictive, medvedeva2021automatic,kaur2019convolutional
Chinese Criminal Courts \cite{luo2017learning,zhong2018legal, yue2021neurjudge, zhong2020iteratively,yang2019legal},
 US Supreme Court \cite{katz2017general,kaufman2019improving}, Indian Supreme Court \cite{malik2021ildc,shaikh2020predicting},
French court of Cassation \cite{csulea2017predicting},
%csulea2017exploring
%Brazilian courts \cite{ bertalan2020predicting},
%lage2022predicting,
Federal Supreme Court of Switzerland \cite{niklaus2021swiss}, 
Turkish Constitutional court \cite{sert2021using}
%,mumcuouglu2021natural},
 UK courts \cite{strickson2020legal} 
 and
 German courts \cite{waltl2017predicting}.
 %,
%the Philippine Supreme court \cite{virtucio2018predicting}, and the Thailand Supreme Court \cite{kowsrihawat2018predicting}

In this study, we focus on classifying case outcomes in the ECHR A and B benchmark tasks introduced by LexGLUE \cite{chalkidis2022lexglue}. Task B involves identifying the set of articles of the European Convention of Human Rights alleged to have been violated by the claimant, while Task A aims to classify which of the convention’s articles has been deemed violated by the court. Both tasks utilize the fact description of the case extracted from the published judgement document as input. Early approaches to LJP relied on rule-based methods \cite{segal1984predicting,kort1957predicting,nagel1963applying}, followed by classification techniques using bag-of-words features \cite{aletras2016predicting, csulea2017exploring}. Recent advancements involve deep learning \cite{zhong2018legal, zhong2020iteratively, yang2019legal} with adoption of pre-trained transformer models \cite{chalkidis2019neural,niklaus2021swiss}, including legal-domain-specific variants \cite{zheng2021does,chalkidis2023lexfiles}. Furthermore, various strategies are explored, such as leveraging dependencies between auxiliary tasks \cite{tyss2023leveraging,yue2021neurjudge,valvoda2023role} or incorporating constraints like contrastive learning \cite{tyss2023leveraging,zhang2023contrastive} and injecting legal knowledge \cite{liu2023ml,tyss2023zero}.

Drawing inspiration from the legal doctrine of \emph{stare decisis}, where precedents (prior cases decided in courts of law) are pivotal in common law jurisdictions to support arguments to arrive at the final outcome. Even in civil law systems, though the prior cases are not directly involved in the final judgment, they are still crucial references during the decision-making process to ensure consistency and proper application of law. We explore how to leverage precedents in LJP models to predict the outcome of a query case. 
While previous research has focused on using precdents for enhancing case representations through contrastive learning \cite{tyss2023leveraging,zhang2023contrastive,gan2022exploiting}, our work investigates two strategies for integrating precedents: (i) direct incorporation at inference through interpolation and (ii) integration during the training process via a fusion layer, enabling the model to reason with similar cases to derive the outcome of the query case, unlike prior works which use similar cases in the form of exemplars in in-context learning to facilitate LLMs in a zero-shot setting \cite{wu2023precedent,shui2023comprehensive}. 

%We deal with how to utilize precedents in LJP models to arrive at outcome of a query case. Such use of similar cases were carried out in earlier studies of LJP by use of contrastive learning where similar case representations are pulled closer in embedding space  but they are confined to learning better case representations. Unlike them, in this work we investigate two strategies to integrate them - (i) directly at inference via interpolation and (ii) even in the training process through a fusion layer facilitating model to learn how to conduct reasoning with the help of them to derive the outcome of query case, 

To enhance relevant precedent retrieval in both strategies, we train a retriever with a fine-grained relevance signal based on the overlap ratio of alleged articles between two cases. For direct integration at inference, we perform label distribution interpolation across each article with the retrieved cases' outcomes, determined by their proximity to the query case. This interpolation has been widely explored in retrieval-augmented KNN-based language models \cite{khandelwal2019generalization,xu2023nearest,he2021efficient}, providing memorization capabilities for rare patterns that are otherwise challenging to capture within a parametric model.  

Additionally, LJP models struggle to effectively leverage retrieved precedents as they are not explicitly trained to relate the query case to retrieved documents and conduct reasoning based on them. Moreover, they focus on memorization during training rather than offloading this process to the retrieval component. To address this, we introduce a precedent fusion module that incorporates precedents during training via stacked cross-attention. Further, we propose joint training that optimizes the retriever model alongside the LJP model, utilizing the relevance signal derived from the fusion module, helping to overcome the latent space divergence issue with static retrievers \cite{izacard2021distilling,izacard2023atlas}. Our experiments demonstrate that integrating precedents at training time along with joint training outperform models with out precedents and with precedents at inference only and without joint training, with larger improvements for sparser articles.

%% file: text/Method.tex
We describe our baseline model which takes the case fact description $x$ and outputs the set of articles as multi-hot vectors (alleged ones in case of Task B and violated ones in Task A). Then we outline the retriever to obtain precedent cases based on the current case. We then introduce two strategies for incorporating information from precedent cases: one during inference and the other during training.

\subsection{Baseline Model}
%Given that case fact descriptions often exceed the maximum input length of a standard BERT-based encoder, 
We adopt a hierarchical model as as outlined in \cite{chalkidis2022lexglue} to account for longer case fact inputs. Each paragraph in the case facts is independently encoded using LegalBERT \cite{chalkidis2020legal}, based on the [CLS] representation. These paragraph representations
%, along with position embeddings, 
are passed through a two-layer transformer to obtain contextualized representations for each paragraph which are then max-pooled to derive the final case representation. This is inputted into a classification layer to produce the multi-hot outcome vector.

\subsection{Precedent Retrieval}
We aim to retrieve legal precedents sharing semantically similar facts with the query case to provide additional supervision for outcome prediction. Due to the lengthy legal documents involved, using a standard retriever isn't feasible. Instead, we adopt a hierarchical architecture akin to the baseline model without the classification head, as our retriever $h$.
%We cannot employ an off the shelf general retriever due to longer documents in our case and thus use the same hierarchical baseline model trained for this task without the classification head as our precedent retriever. 
We employ a pair-wise similarity loss to train the retriever wherein we obtain each case representation through retriever and similarity is computed as the dot product between them. Golden similarity is determined at a fine-grained level using the label overlap ratio (LOR), computed via Jaccard similarity, between the allegation labels of the cases. We compute similarity over allegation labels because they aid in retrieving precedents to tackle challenging negative instances of violation task where the article was alleged but found not to be violated. The loss function is expressed as:
\begin{equation*}
\resizebox{0.8\linewidth}{!}{$
    \begin{aligned}
   L(\theta) &= \text{MSE}\left( (h(x_i) \cdot h(x_j)), \text{LOR}(y_i,  y_j)\right) \nonumber \\
   & \text{LOR}(y_i, y_j) = \frac{|y_i \cap y_j|}{ |y_i \cup y_j|}
\end{aligned}
$}
\end{equation*}

where MSE indicates mean-squared error, $y_{i/j}$ indicate allegation multi-hot  vector of cases $x_{i/j}$ respectively. This approach resembles contrastive learning, making cases with similar allegations lie closer in embedding space \cite{khosla2020supervised,santosh2023leveraging} but at a fine-grained level.

\subsection{Precedents at Inference}
% To enhance our model's predictions directly during inference, we leverage a datastore of prior cases and 

% \noindent \emph{Datastore Construction} 
We construct a datastore $\{K, V\}$ comprising all precedent case representations as keys paired with their multi-hot outcome vectors as values. %$l_i$ derived from the training cases $D$. Formally:
% \begin{equation}
% \{K,V\} = {(h(x_i),l_i)| \forall (x_i, l_i) \in D}
% \end{equation}
%\noindent \emph{Interpolation} 
We retrieve the k-most similar precedent cases $N$ to the query case and incorporate them via label interpolation in a non-parametric way. Given the multi-label classification nature of the task, we interpolate each article separately as a binary classification problem by deriving a  probability estimate for each outcome under each article, considering both the probability assigned to that label and its complement (1 - probability). %To achieve this, we query the datastore using the case representation to find the k-nearest neighbors $N$ based on Euclidean distance. 
Then the distribution of each label under $p_\text{kNN}$ is derived using softmax of their negative distances indicating the closer a precedent case is to the query case, the larger its influence is. 
\begin{align*}
\begin{aligned}
\resizebox{0.88\linewidth}{!}{$
    p_\text{kNN}(l_{ij}|x, x_i) \propto \sum_{(k,v)\in N} \mathbb{1}_{l_{ij} = v_j} \exp\left( \frac{-d(h(x_i), k)}{\tau}\right)
$}
\end{aligned}
\end{align*}

$\tau$ denotes the temperature hyperparameter and d(.) indicates euclidean distance, j denotes the specific article. %$k,v$ denote case representation and respective outcome vector. 
Finally, we interpolate the $p_\text{baseline}$ with $p_\text{kNN}$ to obtain final as:
\begin{equation*}
    \resizebox{0.92\linewidth}{!}{$
    p_{\text{final}}(l_{ij}|x, x_i) = \lambda p_{\text{baseline}}(l_{ij}|x, x_i) + (1 -\lambda) p_{\text{kNN}}(l_{ij}|x, x_i)
    $}
\end{equation*}

\noindent where $\lambda$ serves to balance $p_\text{kNN}$ and $p_\text{baseline}$. 

\subsection{Precedents during training}
We introduce a precedent fusion module to effectively incorporate precedent information during training, separating knowledge memorization from reasoning. This allows the model to focus on understanding the query case and conducting reasoning based on retrieved precedents, rather than solely on memorization. Further, we employ a joint training to optimize both the retriever and LJP models, preventing divergence in the latent space between these two modules if optimized independently.

Upon retrieving precedent cases, we fuse this information into the query using a stacked cross-attention module. This module computes cross-attention between the query case and the retrieved case representations (i.e keys) from the datastore, determining importance weights for their outcome vectors (i.e values). The weighted outcome vectors are transformed into the input space through a feed-forward layer and added to the query representation via a residual connection. This is represented as:
%Once we retrieve precedent cases %from the datastore ${K,V}$ based on the query representation $h_i$ from the LJP encoder
%, we fuse this information into the query using a stacked cross-attention module. Specifically, we compute cross-attention between the query case and the retrieved case representations (i.e., keys of the datastore) to learn the importance weights for their outcome vectors (i.e., the values of the datastore). %based on the similarity of each retrieved case with respect to the query case. 
%This passed through a feed-forward layer to transform the weighted outcome vectors into the input space and add them to the query representation via a residual connection. This cross-attention module is stacked multiple times and this process is represented as:
\begin{equation*}
\resizebox{0.88\linewidth}{!}{$
    h_i^L =h_i^{L-1} + \text{g}( \softmax( \frac{h_i^{L-1}W_q\cdot KW_k}{\sqrt{d_k}}) VW_v )
    $}
\end{equation*}
where $h_i^L$ represents the fused query representation at layer $L$, $\text{g}(\cdot)$ denotes the feed-forward layer, $W_q$, $W_k$, $W_v$ are learnable parameters and $d_k$ represents the dimensionality of the keys representation.

% We train the retriever model along with the LJP model by minimizing the KL-divergence between the retriever similarity scores ($s$) and the cross-attention scores in the fusion module aggregated across layers to obtain a single score for each retrieved precedent ($a$), inspired from \citealt{izacard2021distilling}. The underlying hypothesis is that the cross-attention scores between the query and the precedent cases from the fusion model can be used as a proxy of the similarity for training a better retriever for LJP.

We jointly train the retriever and LJP model by minimizing the KL-divergence between the retriever similarity scores ($s$) and the cross-attention scores in the fusion module, aggregating them across layers to obtain a single score for each retrieved precedent ($a$). This approach, inspired by \citealt{izacard2021distilling}, leverages cross-attention scores as a proxy for similarity to improve retriever tailored in conjunction for the task of LJP.
\begin{equation*}
\resizebox{0.45\linewidth}{!}{$
    L(\theta) = \sum_{k=1}^K a_k \log \frac{a_k}{s_k} 
    $}
\end{equation*}
where $K$ is the number of retrieved precedents. We only optimize retriever parameters with above loss and not the LJP encoder. To account for computational overhead for updating the datastore after every update to the retriever, We allow it to be stale and asynchronously update at certain frequency.

%% file: text/Experiments.tex
% \subsection{Precedent Retrieval Variant}
% We need precedents relevant 

% we want to do so based on document representations relevant to the classification task at hand. A vanilla classifier by design will represent documents in a task-specific manner, i.e. with relevance to the task labels. So we train a vanilla classifier in a first (pre- liminary) phase of training, and use its encoder to obtain representations for all training documents, caching them in a static retrieval repository. 

\input{text/results}

\subsection{Dataset \& Metrics}
We experiment on the ECHR task A and B of LexGLUE \cite{chalkidis2022lexglue}, which consist of 11k case fact descriptions chronologically split into training (2001–2016, 9k cases), validation (2016–2017, 1k cases), and test sets (2017-2019, 1k cases). Both tasks include 10 prominent articles as labels. Following \citealt{chalkidis2022lexglue}, we report micro-F1 and macro-F1 (Mic-/Mac-F1) for both the tasks. We also report hard-macro-F1 (H.Ma-F1) for Task A following \citealt{santosh2022deconfounding}, which is the mean F1-score computed for each article where cases with that article having been violated are considered as positive instances, and cases with that article being alleged but not found to have been violated as negative instances. We also report label overlap ratio (LOR) based on allegation labels for the models with retriever component. Implementation details are described in App. \ref{impl_det}. 

\subsection{Results}
\noindent \textbf{Precedents at Inference:} We investigate the impact of various retriever models by incorporating the precedents retrieved by them directly at inference using label interpolation. We create self-retrieval method that uses the trained LJP model without its classification head as the retriever to obtain precedents. We create a retriever trained with binary relevance loss indicated by at least one shared alleged article, as an alternative to the fine-grained label overlap loss described earlier. Overall, as shown in Table \ref{result_tab}, we observe that adding precedents directly at inference brings improvements in macro-F1 scores for both tasks compared to the baseline, regardless of the retriever used. This suggests that the retrieved precedents, when interpolated at inference, help the model perform better on sparse classes, which may have been difficult to capture with the implicit parameters of the models due to label imbalance in the training data.  Among the retrievers, we find that training with outcome-based relevance signal improves performance compared to using self-retrieval, indicating sub-optimal representations learned by the model through the LJP task alone, which are enhanced by operating directly on the embedding space. Furthermore, the addition of fine-grained relevance loss helps the model learn representations more effectively, as evidenced by the increased performance in downstream tasks and label overlap ratio (LOR). Consequently, we utilize the fine-grained relevance-based retriever for our subsequent experiments.

\noindent \textbf{Fusing Precedents at Training:} We explore different methods for integrating precedent information into the query representation using pre-trained LJP encoder and retriever models, keeping them frozen to isolate the fusion's effect. We devise three fusion variants: (i) computing the mean of retrieved precedent outcomes (values) and adding them to the original representation via a projection layer, (ii) employing cross-attention between the current case representation as the query and the retrieved representation, with labels as key and value vectors and (iii) utilizing stacked cross-attention with multiple layers. Our findings indicate that cross-attention outperforms simple mean, suggesting that the retrieved memory values are not equally important and require learning weights using a  cross-attention. Moreover, learning the contribution of each retrieved precedent also acts as a filter for noisy, non-related precedents provided by the retriever. Furthermore, stacking these cross-attention layers enhances macro-F1 scores across both tasks, underscoring the necessity of complex interactions to learn similarity between the query and retrieved cases. Hence we adopt stacked cross-attention layers for our subsequent experiments.

\noindent \textbf{Training Retriever and LJP:} We unfreeze each of the components to create three variants: training only the LJP model, training both modules jointly and our joint training with additional retriever-specific KL-Divergence loss. Surprisingly, training the LJP model alone leads to a drop in performance for both tasks, primarily reflected in lower LOR values compared to freezing both modules. This suggests that the latent space of the LJP diverges from the frozen retriever, making it challenging to capture relevance. Training the retriever along with the LJP brings performance back to a comparable level to the frozen models but still slightly lags behind, indicating that training the retriever with LJP-specific loss alone does not provide a strong enough signal to learn relevance. Overall, the attention-score-based KL-divergence serves as a better proxy to guide the retriever to follow the latent space of the LJP model to provide better precedents (as seen in higher LOR), resulting in improved downstream performance across both tasks.

%% file: text/results.tex
% \usepackage{multirow}
\begin{table*}[]
\centering
\scalebox{0.8}{
\begin{tabular}{|ll|ccc|cccc|}
\hline
\multicolumn{2}{|l|}{\textbf{}} & \multicolumn{3}{c|}{\textbf{Task B}} & \multicolumn{4}{c|}{\textbf{Task A}} \\ \hline
\multicolumn{1}{|l|}{\textbf{Model}} &
\multicolumn{1}{|l|}{\textbf{Retriever}} &
\multicolumn{1}{c|}{\textbf{Mac-F1}} & \multicolumn{1}{c|}{\textbf{Mic-F1}} & \textbf{LOR} & \multicolumn{1}{c|}{\textbf{H.Ma-F1}} & \multicolumn{1}{c|}{\textbf{Mac-F1}} & \multicolumn{1}{c|}{\textbf{Mic-F1}} & \textbf{LOR} \\ \hline
\multicolumn{1}{|l|}{Baseline}  &  \multicolumn{1}{|l|}{-}    & \multicolumn{1}{c|}{73.56}  & \multicolumn{1}{c|}{79.45}  & -    & \multicolumn{1}{c|}{62.02}   & \multicolumn{1}{c|}{64.18}  & \multicolumn{1}{c|}{70.42}  & -    \\ \hline
\multicolumn{1}{|l|}{\multirow{3}{*}{Inference}}                                                           & Self-Retrieval                                                              & \multicolumn{1}{c|}{74.12}  & \multicolumn{1}{c|}{79.88}  & 0.58 & \multicolumn{1}{c|}{62.83}   & \multicolumn{1}{c|}{64.78}  & \multicolumn{1}{c|}{69.17}  & 0.54 \\ 
\multicolumn{1}{|l|}{}                                                                                     & Binary Rel.                                                                 & \multicolumn{1}{c|}{74.94}  & \multicolumn{1}{c|}{79.77}  & 0.61 & \multicolumn{1}{c|}{63.77}   & \multicolumn{1}{c|}{65.02}  & \multicolumn{1}{c|}{\emph{70.16}}  & 0.58 \\ 
\multicolumn{1}{|l|}{}                                                                                     & Fine-grained Rel.                                                           & \multicolumn{1}{c|}{\emph{75.63}}  & \multicolumn{1}{c|}{\emph{80.28}}  & \emph{0.65} & \multicolumn{1}{c|}{\emph{65.14}}   & \multicolumn{1}{c|}{\emph{66.84}}  & \multicolumn{1}{c|}{69.81}  & \emph{0.63} \\ \hline
\multicolumn{1}{|l|}{\multirow{3}{*}{\begin{tabular}[c]{@{}l@{}}Freeze LJP \\ and Retreiver\end{tabular}}} & Mean                                                               & \multicolumn{1}{c|}{75.48}  & \multicolumn{1}{c|}{79.92}  & 0.65 & \multicolumn{1}{c|}{65.81}   & \multicolumn{1}{c|}{67.25}  & \multicolumn{1}{c|}{71.32}  & 0.63 \\ 
\multicolumn{1}{|l|}{}                                                                                     & Cross Attention                                                             & \multicolumn{1}{c|}{75.81}  & \multicolumn{1}{c|}{\emph{80.98}}  & 0.65 & \multicolumn{1}{c|}{66.23}   & \multicolumn{1}{c|}{68.19}  & \multicolumn{1}{c|}{71.26}  & 0.63 \\ 
\multicolumn{1}{|l|}{}                                                                                     & Stack Cross Att.                                                            & \multicolumn{1}{c|}{\emph{76.61}}  & \multicolumn{1}{c|}{79.62}  & 0.65 & \multicolumn{1}{c|}{67.12}   & \multicolumn{1}{c|}{\emph{68.66}}  & \multicolumn{1}{c|}{\emph{\textbf{72.17}}}  & 0.63 \\ \hline
\multicolumn{1}{|l|}{Train LJP only}                                                                            & \multirow{3}{*}{\begin{tabular}[c]{@{}l@{}}Stack\\ Cross Att.\end{tabular}} & \multicolumn{1}{c|}{74.73}  & \multicolumn{1}{c|}{78.15}  & 0.62 & \multicolumn{1}{c|}{64.88}   & \multicolumn{1}{c|}{65.69}  & \multicolumn{1}{c|}{70.88}  & 0.59 \\
\multicolumn{1}{|l|}{Train both}                                                                           &                                                                             & \multicolumn{1}{c|}{76.45}  & \multicolumn{1}{c|}{80.47}  & 0.66 & \multicolumn{1}{c|}{66.77}   & \multicolumn{1}{c|}{68.29}  & \multicolumn{1}{c|}{71.65}  & 0.62 \\ 
\multicolumn{1}{|l|}{Train both with KLD.}                                                                &                                                                             & \multicolumn{1}{c|}{\emph{\textbf{77.82}}}  & \multicolumn{1}{c|}{\emph{\textbf{81.29}}}  & \emph{\textbf{0.70}} & \multicolumn{1}{c|}{\emph{\textbf{67.79}}}   & \multicolumn{1}{c|}{\textbf{\emph{69.12}}}  & \multicolumn{1}{c|}{\emph{71.74}}  & \emph{\textbf{0.66}} \\ \hline
\end{tabular}}
\caption{Results on Task A and Task B. Rel., Att., KLD.  indicate relevance, attention and KL-Divergence respectively. Best results overall and in each group are bolded and italicized respectively.}
\label{result_tab}
\end{table*}

%% file: text/conclusion.tex
We enhance ECHR outcome classification by integrating precedents into LJP models during training. This improvement is driven by three components: (i) an effective retriever trained with fine-grained relevance signals of allegation labels, (ii) precedent fusion models enabling the offloading of memorization to the retrieval step and reasoning with cues from precedents and (iii) joint training of the retriever alongside the LJP model, enhancing its representations. These precedents also equips models to provide explanations through analogous case-based reasoning, warranting further investigation.

%% file: text/limitations.tex
It's important to acknowledge that despite being labeled as ``legal judgment prediction'' tasks, the fact statements are typically not finalized until the decision outcome is known. This characteristic transforms the task into one of retrospective classification rather than prediction \cite{medvedeva2021automatic}. Although this introduces distracting and confounding phenomena, as highlighted in \citealt{santosh2022deconfounding}, the dataset remains valuable for developing NLP models that analyze fact statements for text patterns corresponding to specific convention articles drafted by the court.

In our study, we demonstrated enhancements in outcome classification performance for ECtHR cases through the incorporation of precedents. While our techniques for precedent integration and retriever training objectives are generalizable and applicable to any jurisdiction, it's important to note that the experimental findings are specific to the context of the ECHR court. The degree of improvement achieved may depend on patterns of drafting texts that may form the basis for computing similarities to retrieve relevant precedents, which in turn can affect downstream performance.

It's worth mentioning that one can directly employ cases cited in the reasoning section of the documents to train retrievers or to validate the effectiveness of retrieved precedents. However, we did not use them directly due to the disguised positive problem \cite{santosh2024ecthr}. This problem arises because a case can be cited for various reasons, such as being authoritative or due to familiarity bias of the drafter. Additionally, it's not feasible to cite all relevant prior cases, leaving the possibility of non-cited cases being related in disguise. Hence, we used LOR based on allegation labels as a signal of relevance, which may be weaker and coarse-grained in nature, as two cases with the same allegation labels might have different involving factors, making them weakly relevant. This approach only provides a lower bound of precise relevance. Future works can explore stronger relevance signals to learn and evaluate those precedents.

Furthermore, our proposed precedent incorporation strategies focus solely on the facts and respective outcomes of prior cases. This approach may be sub-optimal compared to the actual scenario where humans utilize the entire case document, including the reasoning section, which involves argumentation to arrive at the outcome by deducing the application of relevant law in the given context.  In future, it would be beneficial to design models that can integrate these argument sections of precedent cases. By incorporating argumentation, models can deduce outcomes in a more explainable manner by learning applicable arguments within the query context, enhancing the transparency and interpretability of the decision-making process.

%% file: text/ethics.tex
Our experiments were conducted on a dataset of ECHR decisions, which is publicly available as part of the LexGLUE benchmark \cite{chalkidis2022lexglue} and sourced from the public court database HUDOC\footnote{\url{https://hudoc.echr.coe.int}}. While these decisions contain real names and are not anonymized, we do not anticipate any harm beyond the disclosure of this information. However, it's important to acknowledge that utilizing historical data to train models may lead to classifiers that exhibit biased behavior. For instance, \citealt{chalkidis2022fairlex} investigated disparities in classification performance concerning an applicant's gender, age, and the respondent state. Additionally, by leveraging pre-trained encoders, our models inherit any biases encoded within them. However, legal NLP systems leveraging case outcome information and intended for practical deployment should naturally be scrutinized against applicable equal treatment imperatives regarding their performance, behavior, and intended use \cite{baumgartner2024towards}. 

The task of LJP raises significant ethical and legal concerns, both in general and specifically within the context of the European Court of Human Rights (ECtHR) \cite{fikfak2021future}. We do not advocate for the practical implementation of LJP/COC systems by courts. As demonstrated by \citealt{santosh2022deconfounding}, these systems rely on superficial, statistically predictive signals that lack legal relevance, highlighting the risks associated with deploying predictive systems in high-stakes domains such as law. They argue that models utilizing case outcome signals must be developed cautiously, aiming to align their inferences with legal expert reasoning. This aligns with the broader legal NLP community's increasing focus on the ethical aspects of developed systems in technical research \cite{medvedeva2021automatic, medvedeva2023rethinking, tsarapatsanis2021ethical, leins2020give}.

In this study, we utilize LJP as a technical benchmarking task for the development and analysis of neural NLP models on legal text. Our primary objective is to make incremental technical advancements towards enabling systems to work with precedents in a manner that mirrors human experts' analysis of case facts through interactions with past cases. We do not advocate for the practical application of LJP systems by courts but rather aim to explore how their core functionality of processing legal text can align with expert practices as closely as possible. Consequently, our results should be interpreted as technical contributions aimed at advancing models capable of deriving insights from legal data in a legally, ethically, and methodically sound manner. Our research group is dedicated to furthering research on such models to promote transparency, accountability, and explainability of data-driven systems in the legal domain.

%% file: text/appendix.tex
\section{Implementation Details}
\label{impl_det}
We set the maximum token sequence length and maximum number of segments in hierarchical models to 128 and 64, respectively. We train the LJP models using the Adam optimizer \cite{kingma2014adam} with an initial learning rate of 3e-5, employing linear decay and 1000 warmup steps for up to 30 epochs. To optimize training efficiency, we utilize mixed precision and gradient accumulation techniques. The retriever module is initialized using the LegalBERT \cite{chalkidis2020legal} model and further pre-trained for 50,000 pairs of cases, which are sampled uniformly across the entire range of relevance scores from 0 to 1. We employ the Faiss library \cite{johnson2019billion} to construct a datastore of precedents, facilitating efficient similarity computations during retrieval. For incorporating precedents at inference time, we perform a grid search over the interpolation factor ($\lambda$) in increments of 0.1 within the range of [0, 1] to select the best model based on the validation set. Additionally, we vary the value of  k over powers of 2 from 8 to 256. In training incorporation via the fusion layer, we set the number of stacked cross-attention layers to 4. The index is refreshed in joint training every epoch, and we set the number of retrieved precedents during training to 7.